# The Supervised IBP: Neighbourhood Preserving Infinite Latent Feature Models

**Novi Quadrianto**
University of Cambridge
Cambridge, UK
novi.quadrianto@eng.cam.ac.uk

**Viktoriia Sharmanska**
IST Austria
Klosterneuburg, Austria
viktoriia.sharmanska@ist.ac.at

**David A. Knowles**
Stanford University
Stanford, CA, US
knowles84@gmail.com

**Zoubin Ghahramani**
University of Cambridge
Cambridge, UK
zoubin@eng.cam.ac.uk

## Abstract

We propose a probabilistic model to infer supervised latent variables in the Hamming space from observed data. Our model allows simultaneous inference of the number of binary latent variables, and their values. The latent variables preserve neighbourhood structure of the data in a sense that objects in the same semantic concept have similar latent values, and objects in different concepts have dissimilar latent values. We formulate the supervised infinite latent variable problem based on an intuitive principle of pulling objects together if they are of the same type, and pushing them apart if they are not. We then combine this principle with a flexible Indian Buffet Process prior on the latent variables. We show that the inferred supervised latent variables can be directly used to perform a nearest neighbour search for the purpose of retrieval. We introduce a new application of dynamically extending hash codes, and show how to effectively couple the structure of the hash codes with continuously growing structure of the neighbourhood preserving infinite latent feature space.

## 1  Introduction

In statistical data analysis, latent variable models are used to represent components or properties of data that have not been directly observed, or to represent hidden causes that explain the observed data. In many cases, a natural representation of an object would allow each object to admit multiple latent features. Classical statistical techniques require the number of latent features to be fixed a priori. Recently, nonparametric Bayesian models have emerged as an elegant approach to deal with this issue by allowing the number of features to be inferred from data. One class of these models utilise the Indian Buffet Process (IBP) prior (Griffiths & Ghahramani, 2005) to allow an unbounded number of features. Almost all IBP-based statistical models are geared towards *unsupervised* latent feature learning. While unsupervised latent feature models are promising, for example, as an exploratory tool for discovering compact hidden structures in observed data, in many practical settings we seek *supervised* latent variables, that are semantically meaningful and encode supervised side information. Such supervised side information can be expressed as neighbours (similar) and non-neighbours (dissimilar) data pairs, as in (Schultz & Joachims, 2003) for example, and can be used for retrieval of semantically similar neighbours (Weiss et al., 2009).

This paper presents a method to simultaneously infer the dimension of the binary latent representations, and their values so as to encode supervised side information. Binary representations are very attractive for reducing storage requirements and accelerating search and retrieval in large collections of high dimensional data. In recent years, there has been a lot of interest in designing compact binary hash codes such that vectors that are similar in the original data space are mapped to similar binary strings as measured by Hamming distance (Salakhutdinov & Hinton, 2007). However, existing hashing work is typically performed in a static way, that is, a fixed number of bits has to be discovered to model data. We aim to have an approach that is flexible to add bits automatically in order to model new unseen data. This is useful for adaption of hash code to the dynamic and streaming nature of the Internet data, for example.

We present an application of our method to dynamic hash code extension in image retrieval. This application combines the merits of *non-Bayesian* methods that can handle large data (such as Weiss et al. (2009)) and *non-parametric Bayes* that allows extension of codes as required. Our goal is to utilise existing binary hash codes, and learn how to *extend* the codes *pro re nata* in a supervised way when more data become available. For a model that attempts to identify hash codes of dynamically changing data, we argue that the assumption on the number of extended bits to be fixed beforehand is unrealistic.

Our supervised latent variable model enforces latent variables associated with objects of the same semantic concept to have similar values, and latent variables associated with objects of different concepts to have dissimilar values. To achieve this, we define a likelihood function in Section 2 that views this criterion as *preference* relation. When coupled with a flexible prior on infinite sparse binary matrices and a data likelihood, we are able to characterise a probabilistic model for supervised infinite latent variables problems. For the data likelihood, we explore two directions: a standard linear Gaussian model, and our proposed linear probit dependent model, detailed in Section 3. We discuss inference in Section 4 and predictive distribution of our model in Section 5. We present experimental results, including an application in extending hash codes, in Section 6, and draw conclusions in Section 7. First, however, we give a short overview of related work to provide some context for our contributions.

### 1.1 Nearest Neighbour Retrieval

The majority of retrieval techniques today rely on some form of nearest neighbour search. Supervision is an integral component to improve the quality of retrieved results. This is achieved, for instance, in a query-dependent manner by analysing pairs of documents that are returned in response to the text query (Schultz & Joachims, 2003). The supervised information is used to perform metric learning, which maps the original representation of the data samples to a new, preferably low-dimensional, representation where similar samples have small Euclidean distance, and dissimilar samples are separated by a large distance.

For datasets with millions or even billions of entries, even approximate nearest neighbours search techniques such as randomised neighbourhood graphs (Arya et al., 1998) and cover trees (Beygelzimer et al., 2006) are typically infeasible, and one has to resort to hashing approaches. Hash code is a short binary string that can act as an index to directly access elements in a database. Indyk & Motwani (1998) introduce locality sensitive hashing, which purely relies on randomisation techniques yet provides guarantees of preserving metric similarity for sufficiently *long* codes. Next, several machine learning methods have been developed to learn a *compact* hash code Salakhutdinov & Hinton (2007); Torralba et al. (2008); Weiss et al. (2009); Norouzi et al. (2012) among others, and to learn hash codes with better discrimination power, Mu et al. (2010); Wang et al. (2012) for example.

### 1.2 Distance Metric Learning

Approximate nearest neighbour search in general, and hashing-based approaches in particular, provide a powerful and well developed tool for efficient nearest neighbour retrieval from large databases. However, they typically rely on the availability of a meaningful Euclidean metric between the data samples. If such a metric is not readily available, metric learning can be applied to construct one. Basic unsupervised techniques in this area are PCA for dimensionality reduction and the suppression of noise, or its non-linear generalisation, kernel-PCA. Supervised techniques typically work by learning linear projections that place related samples closer together, and unrelated samples farther apart. Often they are based on optimising parametric distance functions such as the Mahalanobis distance with a maximum margin criteria (Schultz & Joachims, 2003; Weinberger & Saul, 2009; Quadrianto & Lampert, 2011), or approximately minimising the leave-one-out classification error as in neighbourhood component analysis of Goldberger et al. (2004).

### 1.3 Infinite Latent Feature Models

Another popular approach for discovering low dimensional structure from high dimensional data is based on latent feature models. We are interested in Indian Buffet Process (IBP) based models that allow number of the latent features to be learnt from data. By defining appropriate data generating likelihood functions, the IBP can be used in, among others, binary factor analysis (Griffiths & Ghahramani, 2005), choice behaviour modelling (Görür et al., 2006), sparse factor and independent component analysis (Knowles & Ghahramani, 2007), link prediction (Miller et al., 2009), and invariant features (Austerweil & Griffiths, 2010; Zhai et al., 2012). For a recent comprehensive review of the IBP models, please refer to Griffiths & Ghahramani (2011). Lately, there is also surging interest in making the latent representations dependent on some known degree of relatedness among observed data (Williamson et al., 2010), in learning correlated non-parametric feature models (Doshi-Velez & Ghahramani, 2009; Miller et al., 2008), or in the direction of supervised modelling, as for dimensionality reduction (Rai & Daume III, 2009). In a recent technical report, Gershman et al. (2012) express a goal closely related to ours, that nearby data is more likely to share latent features than distant data (as induced by distances between data in time or space, for example). However, encouraging sharing features between nearby data does not provide sufficient margin of separation between features of distant data. Our goal is to discover a binary latent space where meaningful notions of similarity and difference are preserved in term of metric distances.

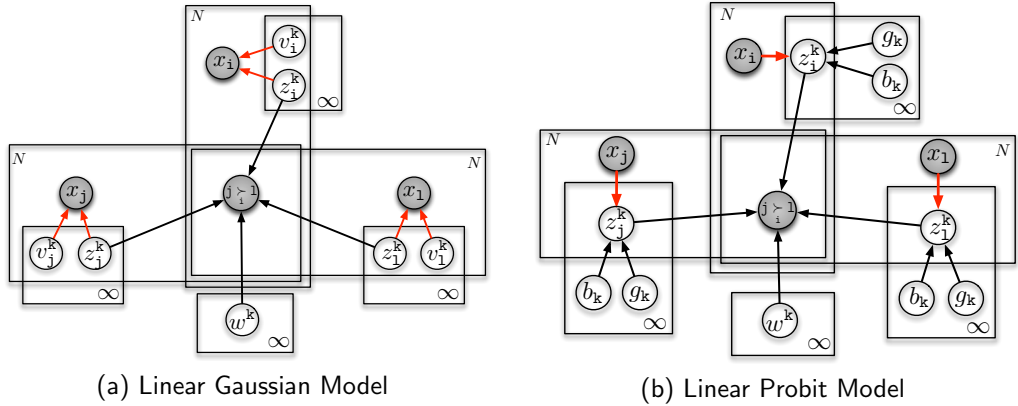

(a) Linear Gaussian Model  (b) Linear Probit Model

Figure 1: Graphical model for our supervised infinite latent variable models based on preference relations. Shade indicates the observed variables. The term $j \succ_i l = 1$ denotes that object $i$ prefers object $j$ to object $l$. 1(a): linear Gaussian model is used to generate the observed data based on latent features. 1(b): linear probit model is used to generate latent features based on the observed data. Note the difference between 1(a) and 1(b) is encoded in the direction of arrows modelling the dependency between data $\mathbf{x}$ and latent features $\mathbf{z}$.

## 2 The Neighbourhood Model

We are given a set of $N$ observed data samples $\{\mathbf{x}_1, \ldots, \mathbf{x}_N\} \subset \mathcal{X}$, and we have used $\mathcal{X}$ to denote the input space. For example, for image objects, $\mathcal{X}$ can be features extracted based on the content of the image itself. We further assume that supervised side information is available in the form of *triplet set* $\mathcal{T} = \{(i, j, l) : i \sim j, i \not\sim l\}$, such that sample $i$ has similar semantic concept to sample $j$ and has no similar concept to sample $l$. Metaphorically, samples $i$ and $j$ are considered neighbours whereas samples $i$ and $l$ are non-neighbours. Usually this type of supervised similarity triplets do not require explicit class labels and thus are easier to obtain. For instance, in a content based image retrieval, to collect feedback, users may be required to report whether an image $\mathbf{x}_i$ looks more similar to $\mathbf{x}_j$ than it is to a third image $\mathbf{x}_l$. This task is typically much easier in comparison to labelling each individual image.

For each data point $\mathbf{x}_n$, we introduce a $K$ dimensional vector $\mathbf{z}_n$ from a *binary* latent space, where $z_n^k = 1$ denotes that object $n$ possesses feature $k$, and $z_n^k = 0$ otherwise, and $K$ is inferred from data. Targeting directly our goal of learning neighbourhood preserving latent space that is suitable for nearest neighbour search, we require that $\mathbf{z}_i$ is similar to $\mathbf{z}_j$ to model $i \sim j$, and $\mathbf{z}_i$ is dissimilar to $\mathbf{z}_l$ to model $i \not\sim l$. The underlying idea of learning the supervised representations is based on a *folk-wisdom* principle (Goldberger et al., 2004; Weinberger & Saul, 2009; Quadrianto & Lampert, 2011) of pulling objects together if they are similar, and pushing them apart if they are not. Further, this principle is formalised as a preference relation.

When we observe that objects $i$ and $j$ are neighbours while objects $i$ and $l$ are non-neighbours, we say that object $i$ prefers object $j$ to object $l$, and use a notation $j \succ_i l$. Let $\mathbf{T}$ be an $N \times N \times N$ preference tensor with entries $\{t_{jl}^i\}$ where $t_{jl}^i = 1$ whenever $j \succ_i l$ is observed. Let $\mathbf{w}$ be a $K \times 1$ *non-negative* weight vector that affects the probability of preference relations among object $i$, $j$, and $l$. We assume that preference relations are independent conditioned on $\mathbf{Z}$ and $\mathbf{w}$, and furthermore only the latent features of objects $i$, $j$, and $l$ influence the tendency of $i$ preferring $j$ to $l$. With the above assumptions, the label preference likelihood function is given by

$$\Pr(\mathbf{T}|\mathbf{Z}, \mathbf{w}) = \prod_{(i,j,l) \in \mathcal{T}} \Pr(t_{jl}^i = 1 | \mathbf{z}_i, \mathbf{z}_j, \mathbf{z}_l, \mathbf{w}). \quad (1)$$

We will subsequently use $p_{jl}^i$ to denote $\Pr(t_{jl}^i = 1|\mathbf{z}_i, \mathbf{z}_j, \mathbf{z}_l, \mathbf{w})$. We define the individual preference probability as follows:

$$p_{jl}^i = \frac{1}{C} \sum_k w_k \mathbb{I}[z_i^k = z_j^k](1 - \mathbb{I}[z_i^k = z_l^k]), \quad (2)$$

where $C = \sum_k w_k \mathbb{I}[z_i^k = z_j^k](1 - \mathbb{I}[z_i^k = z_l^k]) + \sum_k w_k(1 - \mathbb{I}[z_i^k = z_j^k])\mathbb{I}[z_i^k = z_l^k]$ is the normalising constant. In the above, we make use of Iverson's bracket notation: $\mathbb{I}[P] = 1$ for the condition $P$ is true and it is 0 otherwise. The term $\sum_k w_k \mathbb{I}[z_i^k = z_j^k](1 - \mathbb{I}[z_i^k = z_l^k])$ collects the weights for all features that object $i$ and $j$ have but object $l$ does not have, and the weights for all features that object $i$ and $j$ do not have, but $l$ has. Thus, the choice between two alternatives $j$ and $l$ from point-of-view $i$ depends on latent features that are shared between $i$ and $j$ but not $l$. This type of preference model (2) is inspired by the choice model of Görür et al. (2006) and is based on a standard Restle's choice model in psychology (Restle, 1961).

We take a fully Bayesian approach by treating latent variables $\mathbf{Z}$ and $\mathbf{w}$, as random variables, and computing the posterior distribution over them by invoking Bayes' theorem. We discuss the selection of prior probabilities on $\mathbf{Z}$ and $\mathbf{w}$ in detail in the next section.

## 3 The Generative Process

We want to define a flexible prior on $\mathbf{Z}$ that allows simultaneous inference of the number of features and all the entries in $\mathbf{Z}$ at the same time. We will thus put the Indian Buffet Process (IBP) prior (Griffiths & Ghahramani, 2005) on $\mathbf{Z}$. The IBP is a prior on infinite binary matrices such that with probability one, a feature matrix drawn from it will only have a finite number of non-zero features for a finite number of samples (entries). More importantly, the IBP prior has a full support for any feature matrix regardless of the number of non-zero features it has. We choose to put a *Gamma* distribution as a prior for the elements of $\mathbf{w}$. This is a natural prior for a non-negative weight vector. Section 2 describes a likelihood function for modelling the supervised side information, the next required modelling is to define the data likelihood. We explore two directions: one is to use a standard linear Gaussian model which assumes data are generated via a linear superposition of latent features. The second one is to make the latent features be dependent on observed data via a novel and simple linear probit model. We discuss both models in the next sections (refer to Figure 1 for graphical model representations).

### 3.1 $\mathbf{Z} \to \mathbf{X}$ Linear Gaussian Feature Model

This data generating model was initially explored for the IBP in the *unsupervised* context by Griffiths & Ghahramani (2005). In this model, for an $M$-dimensional input space $\mathcal{X} = \mathbb{R}^M$, the data point $\mathbf{x}_n \in \mathbb{R}^M$ is generated as follows:

$$\mathbf{x}_n = \mathbf{V}\mathbf{z}_n + \sigma_x \boldsymbol{\epsilon} \text{ where } \boldsymbol{\epsilon} \sim \mathcal{N}(\boldsymbol{\epsilon}|\mathbf{0}, \mathbf{I}). \quad (3)$$

In the above, $\mathbf{V}$ is a real-valued $M \times K$ matrix of weights. We use a spherical Gaussian conjugate prior with a covariance matrix $\sigma_v^2 \mathbf{I}$ for this feature weight matrix, $\mathbf{V}$. The generative process for our preference model with a linear Gaussian likelihood is then:

$$\mathbf{Z} \sim \text{IBP}(\alpha); \ \mathbf{V} \sim \mathcal{N}(\mathbf{0}, \sigma_v^2 \mathbf{I}); \ \mathbf{x}_n | \mathbf{z}_n, \mathbf{V} \sim \mathcal{N}(\mathbf{V}\mathbf{z}_n, \sigma_x^2 \mathbf{I});$$
$$w_k \overset{\text{i.i.d.}}{\sim} \mathcal{G}(\gamma_w, \theta_w); \ j \underset{i}{\succ} l | \mathbf{Z}, \mathbf{w} \sim \text{Bernoulli}(p_{jl}^i),$$

where $p_{jl}^i$ is defined in Equation (2). We can subsequently compute the posterior distribution of the latent feature matrix $\mathbf{Z}$ and the weights $\mathbf{w}$ using the conditional independence assumptions depicted in Figure 1(a). This is given as, $\Pr(\mathbf{Z}, \mathbf{w}|\mathbf{X}, \mathbf{T}) \propto$

$$\int \Pr(\mathbf{T}|\mathbf{Z}, \mathbf{w}) \Pr(\mathbf{X}|\mathbf{Z}, \mathbf{V}, \sigma_x) \Pr(\mathbf{Z}|\alpha) \Pr(\mathbf{V}|\sigma_v) \Pr(\mathbf{w}|\gamma_w, \theta_w) d\mathbf{V}. \quad (4)$$

### 3.2 $\mathbf{X} \to \mathbf{Z}$ Linear Probit Dependent Model

The neighbourhood model with linear Gaussian data likelihood requires the inferred latent features to *explain* supervised similarity in given triplets and to *generate* observed data. Modelling observed data is a hard task by itself. Instead, we can devote the latent features to model supervised similarity triplets and to have an IBP model where the probability of a feature $k$ being on is *dependent* on some object covariate information $\mathbf{x}_n \in \mathbb{R}^M$. To achieve a dependent IBP model, we start with the stick breaking construction of the IBP (Teh et al., 2007):

$$z_n^k | b_k \sim \text{Bernoulli}(b_k); \ b_k := v_k b_{k-1} = \prod_{j=1}^k v_j \quad (5)$$

$$v_j \sim \text{Beta}(\alpha, 1) \text{ and } b_0 = 1. \quad (6)$$

Williamson et al. (2010) observe that a Bernoulli($\beta$) random variable $z$ can be represented as

$$z = \mathbb{I}[u < \Phi^{-1}(\beta|\mu, \sigma^2)] \quad (7)$$
$$u \sim \mathcal{N}(\mu, \sigma^2), \quad (8)$$

where $\Phi(\cdot|\mu, \sigma^2)$ is a Gaussian cumulative distribution function (CDF), and for simplicity we focus on the standard Gaussian CDF, that is $\Phi_{0,1}(\cdot) := \Phi(\cdot|0, 1)$.

Therefore, we propose a simple alternative to dependent model of Williamson et al. (2010) by linearly parameterising the cut off variable $u_n^k$, as follows:

$$z_n^k = \mathbb{I}[u_n^k < \Phi_{0,1}^{-1}(b_k)] \quad (9)$$
$$u_n^k = -\mathbf{x}_n^\top \mathbf{g}_k + \epsilon, \quad (10)$$

where $\mathbf{g}_k \in \mathbb{R}^M$ is a vector of regression coefficients for each feature $k$, and $\epsilon \sim \mathcal{N}(0,1)$. Equivalently we can integrate out $\epsilon$, and write $\Pr(z_n^k = 1|\mathbf{x}_n, \mathbf{g}_k, b_k)$

$$= \int \mathbb{I}[\epsilon < \mathbf{x}_n^\top \mathbf{g}_k + \Phi_{0,1}^{-1}(b_k)]\mathcal{N}(\epsilon)d\epsilon \quad (11)$$
$$= \Phi_{0,1}(\mathbf{x}_n^\top \mathbf{g}_k + \Phi_{0,1}^{-1}(b_k)). \quad (12)$$

The interpretation of the dependent model above is, that whether a feature $k$ is on is given by a *probit regression* model, with decreasing biases $\Phi_{0,1}^{-1}(b_k)$, which will ensure that only finitely many features are used. Note that this scenario of dependence on per object covariates $\mathbf{x}_n$ is not covered by the dependent IBP of Williamson et al. (2010). Their model defines a prior over multiple IBP matrices which (for certain settings of the model) are marginally IBP: a similar statement for our construction is meaningless since we only have one IBP matrix. However, our model does have the property that $\mathbf{Z}$ is IBP distributed conditional on $\mathbf{g}_k = 0$ for all $k$. We use a spherical Gaussian conjugate prior with a covariance matrix $\sigma_g^2 \mathbf{I}$ for the regression coefficient matrix, $[\mathbf{g}_1, \mathbf{g}_2, \ldots, \mathbf{g}_K] := \mathbf{G}$. With the above construction, the generative process for our preference model with linear probit likelihood is then:

$$v_j \sim \text{Beta}(\alpha, 1); \quad b_k = \prod_{j=1}^k v_j; \quad \mathbf{G} \sim \mathcal{N}(\mathbf{0}, \sigma_g^2 \mathbf{I});$$
$$z_n^k | \mathbf{x}, \mathbf{g}, \mathbf{b} \sim \text{Bernoulli}(\Phi_{0,1}(\mathbf{x}_n^\top \mathbf{g}_k + \Phi_{0,1}^{-1}(b_k)));$$
$$w_k \stackrel{\text{i.i.d.}}{\sim} \mathcal{G}(\gamma_w, \theta_w); \quad j \underset{i}{\succ} l | \mathbf{Z}, \mathbf{w} \sim \text{Bernoulli}(p_{jl}^i),$$

The posterior distribution of the latent feature matrix $\mathbf{Z}$, the feature presence probability $\mathbf{b}$, the weights $\mathbf{w}$, and the regression coefficient matrix $\mathbf{G}$ using the conditional independence assumptions depicted in Figure 1(b) is then $\Pr(\mathbf{Z}, \mathbf{b}, \mathbf{w}, \mathbf{G}|\mathbf{X}, \mathbf{T}) \propto$

$$\Pr(\mathbf{T}|\mathbf{Z}, \mathbf{w})\Pr(\mathbf{Z}|\mathbf{X}, \mathbf{G}, \mathbf{b})\Pr(\mathbf{w}|\gamma_w, \theta_w)\Pr(\mathbf{G}|\sigma_g)\Pr(\mathbf{b}|\alpha). \quad (13)$$

## 4 Inference

In the inference phase, the goal is to compute the joint posterior over the latent binary feature matrix $\mathbf{Z}$, the non-negative weights $\mathbf{w}$ and the regression coefficient matrix $\mathbf{G}$ (for the linear probit dependent model) as expressed in (4) and (13). For our proposed model, exact inference is computationally intractable. Thus, we employ a Markov Chain Monte Carlo (MCMC) method to explore the posterior distributions.

### 4.1 Sampling for Linear Gaussian Model

**M-H sampling of Z:** The sampler for the binary feature matrix $\mathbf{Z}$ consists of sampling existing features, proposing new features with corresponding weights, and accepting or rejecting them based on the Metropolis-Hasting (M-H) criterion. We sample each row $\mathbf{z}_n$ one after another. For sampling existing features, we have: $\Pr(z_n^k = 1|\mathbf{X}, \mathbf{T}, \mathbf{w}) \propto$

$$\int m_{-n,k}\Pr(\mathbf{T}|\mathbf{w}, Z_{-n,k}, z_n^k = 1)\Pr(\mathbf{X}|Z_{-n,k}, z_n^k = 1, \mathbf{V})\Pr(\mathbf{V})d\mathbf{V}, \quad (14)$$

where $m_{-n,k}$ denote the number of non-zero entries in column $k$ excluding row $n$. For sampling new features, we simultaneously propose $(K_{\text{new}}, \mathbf{Z}_{\text{new}}, \mathbf{w}_{\text{new}})$ where a number $K_{\text{new}}$ for new features are sampled from the prior Poisson$(\alpha/N)$. We propose $\mathbf{w}_{\text{new}}$ from its Gamma prior. We consider this proposal with a M-H acceptance ratio which reduces to the ratio of the likelihoods (Meeds et al., 2007).

**Slice sampling of w:** We sample each of the non-negative weights that correspond to the non-zero features and drop the weights that correspond to zero features. We use a slice sampling procedure of Neal (2003). Due to our Gaussian assumptions, the real-valued weight matrix $\mathbf{V}$ in (14) can be marginalised analytically (Griffiths & Ghahramani, 2005).

### 4.2 Sampling for Linear Probit Model

We adapt a slice sampling procedure with stick breaking representation of Teh et al. (2007).

**Adaptive rejection sampling of b:** The form of the conditional distribution of $\mathbf{b}$ can be found in Teh et al. (2007), and due to log-concavity of this distribution, Teh et al. (2007) suggest to use adaptive rejection sampling (ARS) (Gilks & Wild, 1992) to draw samples.

**Sampling of Z:** As in Teh et al. (2007), given the auxiliary slice variable, we will only update the latent feature for each observation and each dimension where its feature presence probability is below the slice. The required conditional distributions for our case are: $\Pr(z_n^k = 1|\mathbf{x}_n, \mathbf{T}, \mathbf{w}, \mathbf{g}_k, b_k) \propto$

$$\Phi_{0,1}(\mathbf{x}_n^\top \mathbf{g}_k + \Phi_{0,1}^{-1}(b_k))\Pr(\mathbf{T}|\mathbf{w}, Z_{-n,k}, z_n^k = 1). \quad (15)$$

**Slice sampling of w:** Similar to the linear Gaussian case, we update the non-negative weights using a slice sampling procedure.

**Elliptical slice sampling of G:** We propose to sample each component $\mathbf{g}_k$ of the regression coefficient matrix $[\mathbf{g}_1, \mathbf{g}_2, \ldots, \mathbf{g}_K]$ using elliptical slice sampling (ESS) (Murray et al., 2010), an efficient MCMC procedure for training of tightly coupled latent variables with a Gaussian prior.

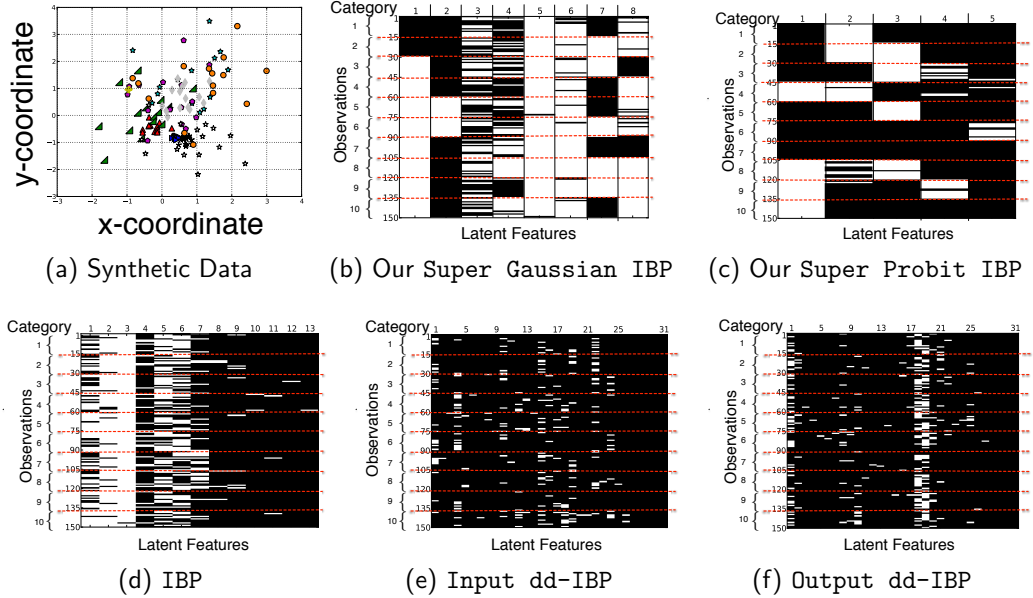

Figure 2: Visualisation of the binary latent space. 2(a): 150 synthetic data points with 10 categories from a mixture of 2-D multivariate Gaussians. 2(b)-2(f): The corresponding binary representations of the data generated by various methods ('1' is white, and '0' is black). The supervised side information is given by a set of triplets. In this example, each training data point has $L$ neighbours from the same category and $L$ non-neighbours from different categories encoded in the form of triplets (*sample*, its *neighbour*, its *non-neighbour*). Here, the binary representations are visualised by grouping observations according to their categories. Our methods preserve given triplets structure by assigning distinct features for different categories.

## 5 Prediction on Test Data

### 5.1 Linear Gaussian Model

For a *previously unseen* test point $\mathbf{x}_* \in \mathbb{R}^M$, the joint predictive distribution for the latent variable $\mathbf{z}_*$ and the preference relation variable $t_*$ is:
$\Pr(\mathbf{z}_*, t_* | \mathbf{X}, \mathbf{T}, \mathbf{x}_*) =$

$$\int \sum_{\mathbf{Z}} \Pr(\mathbf{z}_*, t_* | \mathbf{Z}, \mathbf{w}, \mathbf{X}, \mathbf{x}_*) \Pr(\mathbf{Z}, \mathbf{w} | \mathbf{X}, \mathbf{T}) d\mathbf{w}, \text{ where}$$

$\Pr(\mathbf{z}_*, t_* | \mathbf{Z}, \mathbf{w}, \mathbf{X}, \mathbf{x}_*) = \Pr(t_* | \mathbf{z}_*, \mathbf{w}) \Pr(\mathbf{z}_* | \mathbf{Z}, \mathbf{X}, \mathbf{x}_*)$. This involves averaging over the predictions made by each of the posterior samples of $\mathbf{Z}$ and $\mathbf{w}$. The preference relation variable $t_*$ is a binary variable representing whether object $\mathbf{x}_*$ is preferred or not in some triplet. Since we have trained the binary latent space in a supervised manner, we could predict the neighbours and non-neighbours of the new test point by performing a *nearest neighbour* classification of the inferred test latent variable $\mathbf{z}_*$ with respect to the training latent variables $\mathbf{Z}$. Therefore, we are interested only in the predictive distribution over the latent variable $\mathbf{z}_*$, and it is in the form of:

$$\Pr(\mathbf{z}_* | \mathbf{X}, \mathbf{x}_*) = \sum_{\mathbf{Z}} \Pr(\mathbf{z}_* | \mathbf{Z}, \mathbf{X}, \mathbf{x}_*) \Pr(\mathbf{Z} | \mathbf{X}), \text{ where}$$

$$\Pr(\mathbf{z}_* | \mathbf{Z}, \mathbf{X}, \mathbf{x}_*) \propto \Pr(\mathbf{x}_* | \mathbf{z}_*, \mathbf{Z}, \mathbf{X}) \Pr(\mathbf{z}_* | \mathbf{Z}). \quad (16)$$

We notice that the explicit form of $\Pr(\mathbf{x}_* | \mathbf{z}_*, \mathbf{Z}, \mathbf{X})$ is

$$\begin{bmatrix} \mathbf{X} \\ \mathbf{x}_* \end{bmatrix} \sim \mathcal{N}\left( \mathbf{0}, \begin{bmatrix} \mathbf{Z}\mathbf{Z}^\top + \sigma_x^2/\sigma_v^2 \mathbf{I} & \mathbf{Z}\mathbf{z}_*^\top \\ \mathbf{z}_* \mathbf{Z}^\top & \mathbf{z}_* \mathbf{z}_*^\top + \sigma_x^2/\sigma_v^2 \mathbf{I} \end{bmatrix} \right),$$

thus $\Pr(\mathbf{x}_* | \mathbf{z}_*, \mathbf{Z}, \mathbf{X}) \sim \mathcal{N}(\mu_*, \Sigma_*)$ where

$$\mu_* = \mathbf{z}_*(\mathbf{Z}^\top \mathbf{Z} + \sigma_x^2/\sigma_v^2 \mathbf{I})^{-1} \mathbf{Z}^\top \mathbf{X} \quad (17a)$$

$$\Sigma_* = \mathbf{z}_* \mathbf{z}_*^\top - \mathbf{z}_*(\mathbf{Z}^\top \mathbf{Z} + \sigma_x^2/\sigma_v^2 \mathbf{I})^{-1} \mathbf{Z}^\top \mathbf{Z} \mathbf{z}_*^\top. \quad (17b)$$

The above predictive distribution $\Pr(\mathbf{x}_* | \mathbf{z}_*, \mathbf{Z}, \mathbf{X})$ defines a distribution of the mapping from a latent space to the observed data space.

**Fast approximation** In cases where we are only interested in a maximum a posteriori (MAP) estimate of the latent variables, it is desirable to avoid sampling from the predictive distribution, and directly find an approximate MAP estimate in a computationally efficient way. In our case, we use the predictive mean of $\Pr(\mathbf{x}_* | \mathbf{z}_*, \mathbf{Z}, \mathbf{X})$ in (17a) to approximate $\mathbf{z}_*$ by solving a linear system of equations, resulting in a continuous estimate $\hat{\mathbf{z}}_*$ of the binary vector $\mathbf{z}_*$.

## 5.2 Linear Probit Dependent Model

Similar to the linear Gaussian model but with explicit representation of the regression coefficient matrix, the joint predictive distribution of the latent variable $\mathbf{z}_*$ and the preference variable $t_*$ for a *new* test point $\mathbf{x}_* \in \mathbb{R}^M$ is: $\Pr(\mathbf{z}_*, t_* | \mathbf{T}, \mathbf{x}_*) =$

$$\iiint \Pr(\mathbf{z}_*, t_* | \mathbf{G}, \mathbf{w}, \mathbf{b}, \mathbf{x}_*) \sum_{\mathbf{Z}} \Pr(\mathbf{Z}, \mathbf{b}, \mathbf{w}, \mathbf{G} | \mathbf{X}, \mathbf{T}) d\mathbf{b} d\mathbf{w} d\mathbf{G},$$

with test likelihood given as follows:

$$\Pr(\mathbf{z}_*, t_* | \mathbf{G}, \mathbf{w}, \mathbf{b}, \mathbf{x}_*) = \Pr(t_* | \mathbf{z}_*, \mathbf{w}) \Pr(\mathbf{z}_* | \mathbf{G}, \mathbf{b}, \mathbf{x}_*). \quad (18)$$

As earlier, we are only concerned with the predictive distribution over the latent variable $\mathbf{z}_*$ for the new input $\mathbf{x}_*$, that is $\Pr(\mathbf{z}_* | \mathbf{G}, \mathbf{b}, \mathbf{x}_*)$. Based on our linear probit model, this will simply be $\Pr(\mathbf{z}_*^k = 1 | \mathbf{G}, b_k, \mathbf{x}_*) = \Phi_{0,1}(\mathbf{x}_*^\top \mathbf{g}_k + \Phi_{0,1}^{-1}(b_k))$.

## 6 Experiments

To assess the efficacy of our models, we perform two sets of experiments. We start with a synthetic data experiment to explore the structure of the latent space $\mathbf{Z}$ produced by the proposed models (Section 6.1-6.2). Our second experiment is extending hash codes in image data (Section 6.3).

### 6.1 Visualisation of the Binary Latent Spaces

**Data** We generate 150 synthetic data points with 10 categories from a mixture of 2-D multivariate Gaussians with uniformly drawn standard deviations in the range $[0, 1]$. The means are uniformly drawn in the range $[-1, 1]$ *per category*. The visualisation of the generated data is provided at Figure 2(a).

**Algorithms** We compare the generated latent space of our supervised linear Gaussian (`Super Gaussian IBP`) and supervised linear probit (`Super Probit IBP`) models with the Indian buffet process (`IBP`), and the distance dependent Indian buffet process with distance defined on $\mathcal{X}$ (`Input dd-IBP`), and on the labels (`Output dd-IBP`)[1]. The supervised side information is given by a set of triplets generated the same way as in Weinberger & Saul (2009). Specifically, for each training data point, we are given its $L$ neighbours from the same category, and $L$ non-neighbours from different category encoded in the form of triplets (in this experiment we use $L = 15$). As a practical note, triplets as supervised side information correspond only to a small number of observed entries in $\mathbf{T}$. This translates to a small computational overhead compared to the standard IBP inference. For all methods, we place conjugate priors on the hyperparameters, and subsequently perform posterior inference over them. Our `Super Gaussian IBP` and `Super Probit IBP` methods discover binary representations which preserve neighbourhood structure. The results are shown in Figure 2. In comparison to the IBP 2(d), dd-IBP input 2(e), and dd-IBP output 2(f) models, the inferred feature matrix $\mathbf{Z}$ of the proposed models 2(b) and 2(c) appears to have a notable supervised structure for all categories by assigning distinct features for different categories.

---

[1] We use the implementation provided by Gershman et al. (2012) at http://www.princeton.edu/~sjgershm/ddIBP_release.zip

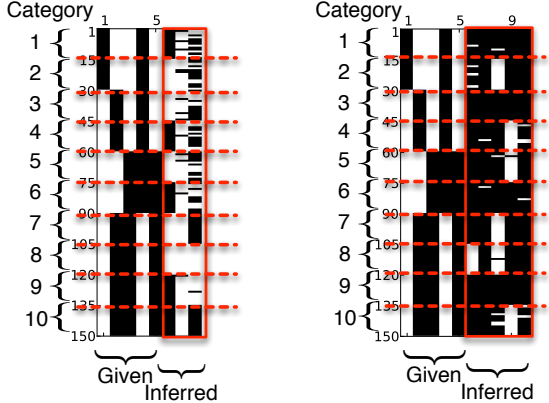

(a) Super Gaussian IBP  (b) Super Probit IBP

Figure 3: Extending observed binary variables. The first 5 *given* binary variables do not respect the neighbourhood structure, for example, categories 1 and 2 have the same '01101' representation. Our supervised models allow *coupling* between given and inferred latent features. As a result, the inferred latent features enforce separation among categories and amend shortcomings that the observed binary representations might have in respecting the neighbourhood structure.

### 6.2 Model with *Observed* Binary Variables

We are interested to explore how our model can be used to extend an existing binary hash of the data. To do so, we assume that we are already given 5 binary variables that partially separates the objects according to their categories. In this case, categories 1 and 2 have binary representations '01101', categories 3 and 4 have '10101' representations, categories 5 and 6 have '11000', and categories 7, 8, 9 and 10 have 10010. Refer to the first 5 dimensions of Figure 3(a) for illustration. The task is to extend the binary vector to model the supervised neighbourhood information, thereby disambiguating classes with the same observed binary hash. In this setting, we want to extend the observed binary representations $\mathbf{h_n} \in \mathcal{H}$ (for each example $\mathbf{x_n}$) where $\mathcal{H} \in \{0,1\}^D$ with a *latent* binary feature $\mathbf{z_n}$, forming an extended representation $[\mathbf{h_n}^\top \mathbf{z_n}^\top]^\top$. Let $\mathbf{H}$ be the observed $N \times D$ binary representation matrix (D = 5 in our experiment) and $\mathbf{w_H}$ be a $D \times 1$ *non-negative* weight vector. The full preference probability is now, $p^i_{jl} = \frac{1}{C} \sum_k w_k \mathbb{I}[z^k_i = z^k_j](1 - \mathbb{I}[z^k_i = z^k_l]) + \sum_d w^d_H \mathbb{I}[h^d_i = h^d_j](1 - \mathbb{I}[h^d_i = h^d_l])$, where the normalising constant $C$ ensures $p^i_{jl} + p^i_{lj} = 1$. The latent features are inferred to enforce separation among categories and amend shortcomings that the observed binary variables might have in respecting the neighbourhood structure.

Table 1: Extending hash codes results for image data. $k$-NN accuracy mean ± std over 5 random repeats. `IBP`: standard IBP algorithm (Griffiths & Ghahramani, 2005); `dd-IBP`: distance dependent IBP (Gershman et al., 2012) where `Input`: distance on $\mathcal{X}$, and `Output`: distance on the labels; `Super Gaussian IBP`: Our proposed supervised IBP with linear Gaussian feature model; `Super Probit IBP`: Our proposed supervised IBP with linear probit dependent model; `Reference`: original 128 real-valued feature representations. The best result and those not significantly worse than it, are in **boldface**. We use a one-sided paired t-test with 95% confidence.

|  | Hash | IBP | Input dd-IBP | Output dd-IBP | Super Gaussian IBP | Super Probit IBP | Reference |
|---|---|---|---|---|---|---|---|
| | | | | 5 animal categories | | | |
| **1 NN** | 26.3±2.2 | 30.3±2.0 | 27.9±6.3 | 29.7±3.5 | 33.8±1.6 | **42.8±2.4** | **40.9±4.7** |
| **3 NN** | | 29.5±2.9 | 29.6±3.6 | 31.0±1.4 | 34.6±2.3 | **41.9±3.4** | **40.2±3.4** |
| **15 NN** | | 31.5±2.6 | 27.8±2.8 | 28.1±3.2 | 35.5±1.0 | **44.5±2.1** | 39.3±3.7 |
| **30 NN** | | 29.5±3.2 | 24.3±3.0 | 23.6±3.4 | 33.8±0.7 | **45.9±4.1** | 36.1±2.8 |
| | | | | 10 animal categories | | | |
| **1 NN** | 12.7±2.5 | 17.1±3.1 | 12.9±2.6 | 15.9±2.3 | 17.3±1.2 | **25.0±2.9** | **25.0±2.2** |
| **3 NN** | | 17.9±2.8 | 13.1±2.4 | 15.3±2.3 | 18.2±1.2 | **25.1±3.0** | **26.0±1.7** |
| **15 NN** | | 16.4±2.5 | 14.7±2.3 | 15.1±1.8 | 18.0±1.5 | **26.6±2.7** | **27.8±2.8** |
| **30 NN** | | 17.7±3.4 | 14.5±1.9 | 14.0±1.9 | 18.3±1.4 | **27.5±2.4** | **25.8±1.4** |

**Results** The supervised models are trained to utilise the *given* binary features, and to add additional binary latent representations only when it is needed to support the discrimination between categories (see Figure 3). As an example, in case of categories 1 and 2 that are indistinguishable under the given 5 binary vectors, 3(a)-3(b) learn at least unit distance in the extended representation for these categories, and increase the separation of the codes for the rest of categories. For this example, `Super Gaussian IBP` (Figure 3(a)) discovers additional 3 binary latent variables where category 1 has '0∗∗' and category 2 has '1∗∗'. While `Super Probit IBP` (Figure 3(b)) discovers 5 more binary latent variables with '∗∗0∗∗' and '∗∗1∗∗' assigned to category 1 and 2, respectively.

### 6.3 Extending Hash Codes Application

In this experiment, we assume that we are given binary hash codes, for example, via a spectral hashing method (Weiss et al., 2009) or via binary attribute predictors (Lampert et al., 2009) explained in the subsequent section, and our goal is to extend these observed codes with latent binary features. We expect the *extended codes* to have a better nearest neighbour search performance, especially in the case where the hash codes do not respect the neighbourhood structure of data.

**Data** We use the Animals with Attributes dataset [2]. The dataset consists of 30,475 images. Each of the images has a category label which corresponds to the animal class. There are 50 animal classes in this dataset. The dataset also contains semantic information in the form of an 85-dimensional Osherson's (Osherson et al., 1991; Kemp et al., 2006) attribute vector for each animal class describing colour, texture, shape, among others. Images are represented by colour histograms of quantised RGB pixels with a codebook of size 128.

**Hash Codes** The hash codes at training phase are given by the Osherson's attribute vector. In the testing phase, we build hash codes using attribute predictors trained offline (Lampert et al., 2009). We generate the hash codes as follows: each class is assigned an attribute binary string of length $D$ (in our case, the Osherson's vector), subsequently we learn $D$ logistic regression functions, one for each bit position in these binary strings. When a new data point arrives, we evaluate each of the $D$ *test* logistic regressors to generate a $D$-bit hash code.

**Results** We use 27,032 images from 45 classes to be our *initial image corpus* for learning the attribute hash codes. We use the colour histograms to represent images, and we focus on colour attributes hash codes, which corresponds to the first 5-bits in the Osherson's attribute vector. This simulates the case where a large pool of data is available for building the hash codes. From the remaining five classes, we randomly sample

---

[2] http://attributes.kyb.tuebingen.mpg.de/

Table 2: Accuracy comparison between SVM with a linear kernel and Super Probit IBP. The best result and those not significantly worse than it, are in **boldface** (one-sided paired t-test with 95% confidence).

|  | 5 categories | 10 categories |
|---|---|---|
| Linear SVM | **43.3±4.0** | **29.0±3.3** |
| Super Probit IBP | **45.9±4.1 (30 NN)** | **27.5±2.4 (30 NN)** |

Table 3: Effect of Bayesian Averaging on Super Probit IBP. Accuracy mean±std. `last`: using a sample from the last iteration; `average`: using samples from the last 50 iterations. **boldface** is significant using a one-sided paired t-test with 95% confidence.

|  | 5 animal categories | | 10 animal categories | |
|---|---|---|---|---|
|  | `last` | `average` | `last` | `average` |
| **1 NN** | 42.8±2.4 | 42.7±3.2 | 25.0±2.9 | 25.5±3.9 |
| **3 NN** | 41.9±3.4 | **44.0±2.7** | 25.1±3.0 | **27.1±2.9** |
| **15 NN** | 44.5±2.1 | **46.5±2.3** | 26.6±2.7 | 27.7±2.3 |
| **30 NN** | 45.9±4.1 | 46.5±2.3 | 27.5±2.4 | 27.2±2.7 |

300 images with uniform class proportions to form a *refinement set* for training our models, and the test set using 50/50 split. Our refinement set simulates the case where training samples are from different categories than in initial corpus, and therefore have different unseen properties. We repeat the above procedure for a refinement set with 10 new categories. That is, we use 23,266 images from 40 classes to learn the hash codes, and randomly sample 600 images from the remaining 10 classes for training and test with 50/50 split. The supervised similarity triplets are formed in the same way as in synthetic experiments ($L = 30$). We note that the costly MCMC procedure is performed offline at the training phase. At test time, we simply perform a fast approximation via matrix vector multiplication in linear Gaussian (Section 5.1) or compute probit regression in linear probit (Section 5.2).

The full results with a comparison to the predicted hash codes using logistic regressors and the standard `IBP` and `dd-IBP` are summarised in Table 1[3]. We observe that our proposed models, `Super Gaussian IBP` and `Super Probit IBP`, exceed the performance of IBP and dd-IBP in *all cases*. We further notice that Super Probit IBP is far superior to Super Gaussian IBP. We credit this to the fact that linear Gaussian models are less suitable for modelling real-valued images (Austerweil & Griffiths, 2010; Zhai et al., 2012). One of the solutions would be to define a more sophisticated likelihood function (Austerweil & Griffiths, 2010; Zhai et al., 2012). Instead in this work we focus on generating binary features that depend on the observed images via probit regression. As a reference, we also provide $k$-NN performance in the original 128 real-valued features. Original features will require storage of 8,192 (128∗64) bits per image, while our Super Probit IBP code with 80 inferred binary latent dimensions will only consume approximately 80 bits per image and gives better results. Further, to put our results in a wider perspective, we also provide results of running SVM[4] on the original real-valued features with a linear kernel in Table 2.

---

[3]The $k$-NN performance of the hash method does not depend on $k$, because for training we use the *given* Osherson's colour hash codes defined per class.

[4]We use the LIBSVM library available at http://www.csie.ntu.edu.tw/~cjlin/libsvm/.

Bayesian approach allows us to learn a distribution over hash codes. In our experiments, we run MCMC until a fixed number of iterations, and subsequently consider the hash codes given by the last iteration. We can instead exploit the full distribution by averaging the nearest neighbour retrieval performances after burn-in period. The results of Bayesian averaging on Super Probit IBP are summarised in Table 3. Clearly, averaging has a *positive effect* in the performance, however, with a price in storage requirement where now several databases have to be maintained.

## 7 Discussion and Conclusion

We have presented probabilistic models to simultaneously infer the number of binary latent variables, and their values so as to preserve a given neighbourhood structure. The models map objects in the same semantic concept to similar latent values, and objects in different concepts to dissimilar latent values. We substantiate our claim that the proposed supervised models encourage coupling among latent features by showing that when given binary representations, the models utilise the given representation, and add dimensions in a latent space when it is needed to preserve the neighbourhood structure.

Our experiments in a nearest neighbour search show that our methods are able to find semantically similar neighbours due to the supervised nature of the latent space, and far exceed the performance of other state-of-the-art infinite latent variable models, such as the standard Indian buffet process (IBP) and its recent extension, the distance dependent IBP (dd-IBP).

## Acknowledgements

The authors would like to thank Kurt Miller, Dilan Görür, and Christoph Lampert for discussions. NQ is supported by a Newton International Fellowship. The work is done while VS and DAK are at University of Cambridge.